%% file: main.tex
\documentclass[runningheads]{llncs}
\usepackage{graphicx}
\usepackage{svg}

\usepackage{color, soul}
\usepackage{amsfonts}
\usepackage{amsmath}
\usepackage{algorithm}
\usepackage{algorithmic}
\usepackage{booktabs}
\usepackage{multirow}
\usepackage{tabularx}
\newcolumntype{Y}{>{\centering\arraybackslash}X} 
\usepackage{amssymb}
\usepackage{pifont}
\newcommand{\xmark}{\ding{55}}%
\usepackage{appendix}
\usepackage[colorlinks=true, linkcolor=black]{hyperref}

\begin{document}
\title{UniVIE: A Unified Label Space Approach to Visual Information Extraction from Form-like Documents}
\titlerunning{UniVIE: A Unified Label Space Approach to VIE}
%
\newcommand*\samethanks[1][\value{footnote}]{\footnotemark[#1]}
\author{Kai Hu\inst{1,2,}\thanks{Work done when Kai Hu and Jiawei Wang were interns, Weihong Lin, Zhuoyao Zhong and Lei Sun were full-time employees in Multi-Modal Interaction Group, Microsoft Research Asia, Beijing, China.} \and
Jiawei Wang\inst{1,2,}\samethanks{} \and
Weihong Lin\inst{2,}\samethanks{} \and \\
Zhuoyao Zhong\inst{2,}\samethanks{} \and
Lei Sun\inst{2,}\samethanks{} \and
Qiang Huo\inst{2}
}
\authorrunning{Hu et al.}
%
\institute{
University of Science and Technology of China, Hefei, China \and
Microsoft Research Asia, Beijing, China \\
\email{hk970213@mail.ustc.edu.cn, wangjiawei@mail.ustc.edu.cn,} \\
\email{lwher1996@outlook.com, zhuoyao.zhong@gmail.com,} \\ 
\email{ray\_ustc@163.com, qianghuo@microsoft.com}}
%


\def \Ours {UniVIE}

\maketitle              
\begin{abstract}
Existing methods for Visual Information Extraction (VIE) from form-like documents typically fragment the process into separate subtasks, such as key information extraction, key-value pair extraction, and choice group extraction. However, these approaches often overlook the hierarchical structure of form documents, including hierarchical key-value pairs and hierarchical choice groups.
To address these limitations, we present a new perspective, reframing VIE as a relation prediction problem and unifying labels of different tasks into a single label space. This unified approach allows for the definition of various relation types and effectively tackles hierarchical relationships in form-like documents. 
In line with this perspective, we present \Ours{}, a unified model that addresses the VIE problem comprehensively. \Ours{} functions using a coarse-to-fine strategy. It initially generates tree proposals through a tree proposal network, which are subsequently refined into hierarchical trees by a relation decoder module. To enhance the relation prediction capabilities of \Ours{}, we incorporate two novel tree constraints into the relation decoder: a tree attention mask and a tree level embedding.
Extensive experimental evaluations on both our in-house dataset HierForms and a publicly available dataset SIBR, substantiate that our method achieves state-of-the-art results, underscoring the effectiveness and potential of our unified approach in advancing the field of VIE.

\end{abstract}
\input{chapters/introduction.tex}
\input{chapters/related_works.tex}
\input{chapters/methodology.tex}

\input{chapters/experiments.tex}
\input{chapters/conclusion.tex}
%
%
%
\bibliographystyle{splncs04}
\bibliography{bibfile}

\end{document}

%% file: chapters/introduction.tex
\section{Introduction}
Form-like documents, encompassing a diverse range of document types, are crucial to various sectors, including finance, healthcare, administration, etc. As a crucial task in the domain of form understanding, Visual Information Extraction (VIE) from form-like documents aims to convert these semi-structured documents into machine-readable formats. This transformation is a vital step in the process of automating and streamlining data processing workflows. However, the inherent complexities such as intricate layouts, hierarchical structures, and diverse semantic interpretations within these documents present significant challenges to the successful execution of VIE.

\begin{figure}[t]
    \centering
    \includegraphics[width=1.0\textwidth]{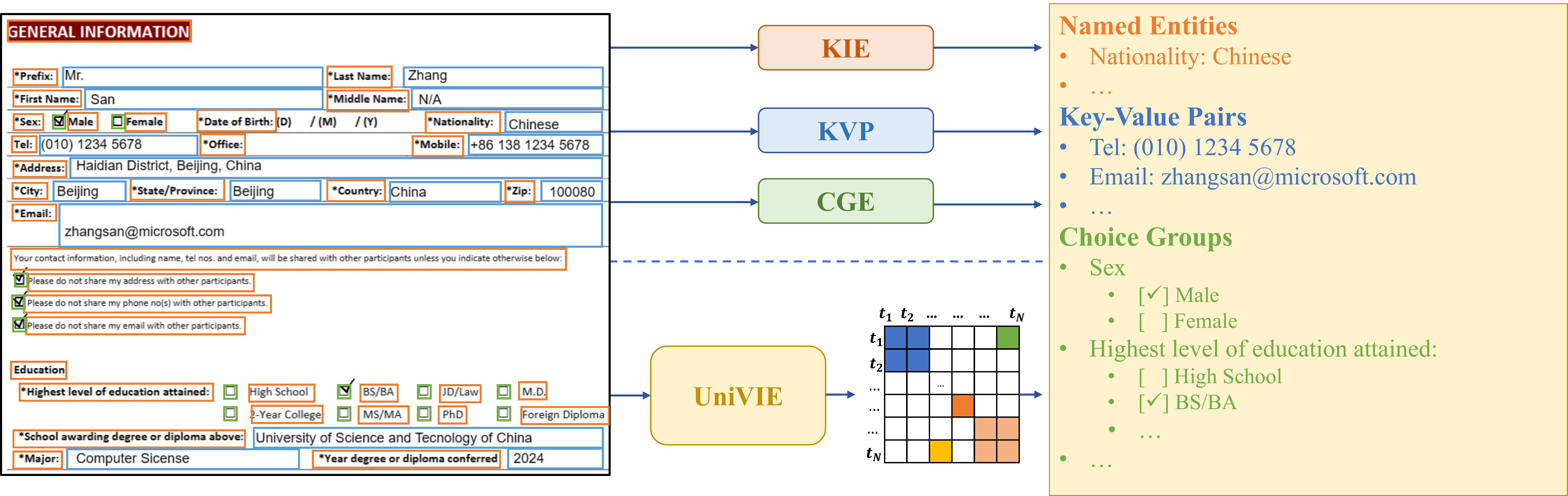}
    \caption{An illustrative example of Visual Information Extraction using our proposed \Ours{} model. (Orange rectangles represent text-lines, green rectangles represent choice widgets, and blue rectangles represent text widgets. Best viewed in color.)}
    \label{fig:example}
\end{figure}

Recently, VIE has garnered considerable interest from both Computer Vision \cite{wang-etal-2022-lilt,zhang2020trie} and Natural Language Processing \cite{xu2020layoutlm,xu2020layoutlmv2} communities, marking significant advancements over traditional rule-based \cite{watanabe1995layout} and template-matching \cite{rastogi2020information} methods. However, existing VIE methods typically fragment the task into several subtasks, including Key Information Extraction \cite{huang2019icdar2019,park2019cord,vsimsa2023docile} (KIE, aka Entity Extraction), Key-Value Pair Extraction \cite{jaume2019funsd,xu2021layoutxlm} (KVP, aka Entity Linking), and Choice Group Extraction \cite{aggarwal2020multi,aggarwal2021form2seq,mathur2023layerdoc} (CGE). Corresponding models are then designed or fine-tuned for each subtask, as illustrated in the upper part of Fig.~\ref{fig:example}. Despite achieving notable results on individual subtasks, these methods often neglect the hierarchical structure inherent in key-value pairs and choice groups, resulting in incomplete final outputs.

Inspired by a table filling strategy \cite{wang2021unire,yan2022utc} utilized in the domain of joint entity and relation extraction, we propose a new perspective: reframing VIE as a relation prediction problem and unifying the labels of different subtasks into a single label space. As depicted in Fig.~\ref{fig:example}, within a given form document image, we identify three fundamental form elements: text-line, text widget, and choice widget. We assert that these basic units are given and our primary task is to predict the hierarchical relationships among these units to extract structured information such as named entities, key-value pairs, and choice groups. We classify the relationships into two types. The first type, \emph{intra-field} relationship, groups basic units into semantically coherent page objects such as \emph{intra-key} relationships within key fields, \emph{intra-value} relationships within value fields of key-value pairs, and similar relationships within choice fields, and choice group titles of choice groups. The second type, \emph{inter-field} relationship, groups semantically coherent page objects into semantically structured page objects, such as \emph{inter-kvp} relationships within key-value pairs and \emph{inter-cg} relationships within choice groups. In inter-field relationships, the first basic unit of the subject field points to the first basic unit of the object field. Consequently, both intra-field and inter-field relationships exist at the basic unit level and converge within the same label space. Decoding these relationship types enables us to extract named entities, key-value pairs, and choice groups effortlessly, while also restoring their hierarchical structure.

Aligned with this new perspective, we present \Ours{}, a unified model designed to address the VIE problem in a comprehensive manner. Initially, \Ours{} employs a pre-trained language model, such as Bert \cite{devlin2018bert}, for text feature extraction and a visual backbone, such as ResNet \cite{he2016deep}, for image feature extraction. These disparate features are concatenated and fed into a \emph{tree proposal network} to generate tree proposals. To further refine these proposals, we introduce a novel \emph{relation decoder}. This module uses relation proposals as queries and leverages self-attention mechanisms to model the interaction among relation proposals and cross-attention mechanisms to model interactions between relation proposals and basic units. Furthermore, we introduce two new tree constraints to the relation decoder: a \emph{tree attention mask} and a \emph{tree level embedding}. These novel components significantly improve the modeling of interactions within hierarchical structures. Finally, a \emph{relation decoding algorithm} is applied to decode hierarchical choice groups and hierarchical key-value pairs from these relationships, producing the hierarchical results. Empowered by these new techniques, \Ours{} achieves state-of-the-art results on both our in-house dataset HierForms and a publicly available dataset SIBR, underscoring the effectiveness and potential of our unified approach in advancing the field of VIE.

The main contributions of this paper can be summarized as follows:

\begin{itemize}

\item We reframe the VIE as a relation prediction problem, unifying the label space of KIE, KVP, and CGE effectively. 

\item We present \Ours{}, a unified model that integrates multiple tasks of VIE into a single framework, employing a coarse-to-fine strategy for efficient and accurate information extraction from complex form-like documents.

\item We propose a novel relation decoder module complemented by two tree constraints within the decoder: a tree attention mask and a tree level embedding to improve the modeling of interactions within hierarchical structures.
\end{itemize}

%% file: chapters/related_works.tex
\section{Related Work}

\subsection{Visual Information Extraction}
Information extraction from visually-rich document images has been studied for decades. Early works \cite{dengel2002smartfix,cesarini2003analysis,medvet2011probabilistic,esser2012automatic,schuster2013intellix,rusinol2013field} primarily relied on predefined rules or manually designed features within known templates. However, these approaches often failed to perform efficiently with unfamiliar templates, thus limiting their practical applicability.
With the development of deep learning, significant progress has been made in VIE. As previously discussed, the VIE process is divided into several subtasks, including key information extraction, key-value pair extraction, and choice group extraction.

\subsubsection{Key Information Extraction.}
The objective of Key Information Extraction (KIE) is to locate, analyze, and extract key entities (aka named entities) contained in form documents. This process involves identifying and categorizing important information such as names, dates, and numerical data, which are crucial for understanding the content of the form document. Current deep-learning based solutions typically treat KIE as a token classification problem, leveraging diverse deep learning architectures like LayoutLM \cite{xu2020layoutlm}, ViBERTgrid \cite{lin2021vibertgrid}, or TRIE \cite{zhang2020trie} to predict the BIO labels for each document token. Nevertheless, these methods struggle with discontinuous named entities, especially when an entity spans multiple lines with a non-trivial reading order of the lines.

\subsubsection{Key-Value Pair Extraction.}
The aim of Key-Value Pair Extraction (KVP) is to establish connections between different entities by pairing related keys and values, thereby providing a more comprehensive understanding of the information within the form. Existing KVP methods \cite{jaume2019funsd,hwang2020spatial,carbonell2021named,xu2021layoutxlm} typically concatenate the representations of each entity pair and apply a classifier, such as a multilayer perceptron (MLP), to predict if these two entities form a key-value pair. Recently, SimpleDLM \cite{gao2021value} and KVPFormer \cite{hu2023question} reframed KVP as a question-answering problem and employed an encoder-decoder Transformer-based QA model to extract key-value pairs from document images.

\subsubsection{Choice Group Extraction.}
Choice Group Extraction (CGE) \cite{aggarwal2021form2seq,aggarwal2020multi}, introduced by Adobe researchers, aims to extract all choice groups present within form documents. A choice group is fundamentally a collection of boolean fields, often accompanied by an optional title text providing instructions or context. This user interface element facilitates user input by enabling users to select from a predefined set of options to express their preferences or provide specific information. Choice groups prove beneficial across various scenarios, such as market research, consumer behavior studies, and digital marketing. Form2seq \cite{aggarwal2021form2seq} proposed a sequence-to-sequence framework to group lower-level constituent elements (\textit{e.g.}, text-line and widgets) into higher-order constructs like text fields, choice fields and choice groups. LayerDoc \cite{mathur2023layerdoc} recursively groups smaller regions into larger semantic elements in a bottom-up, layer-wise fashion. 

Despite the significant achievements of previous methods on individual subtasks, they have overlooked the hierarchical structure inherent in key-value pairs and choice groups, leading to incomplete final outputs. In this paper, we propose a novel perspective: reframing VIE as a relation prediction problem and integrating labels of different subtasks into a unified label space. This unified approach allows for the definition of various relation types and effectively tackles hierarchical relationships in form-like documents. 

\subsection{Table Filling Strategy}
In the domain of joint entity and relation extraction \cite{wang2020two,zheng2017joint,qiao2022joint}, table filling methods \cite{gupta2016table,wang2020tplinker,wang2021unire,yan2022utc} have gained attention. These methods maintain a table for each named entity and relation, where the elements in the i-th row and j-th column of the table correspond to the i-th and j-th words in the input sentence. The main diagonal elements in the table represent entity labels, while the off-diagonal elements usually denote the relationship between two entities. Consequently, the joint entity and relation extraction task is transformed into the task of filling these tables accurately and effectively. The filled table can then be decoded to obtain the final results of named entities and entity relationships. Inspired by table-filling methods, in this paper, we propose a unified label space for key information extraction, key-value pair extraction, and choice group extraction.

%% file: chapters/methodology.tex
\section{Problem Definition}

\begin{figure}[t]
    \centering
    \includegraphics[width=0.8\textwidth]{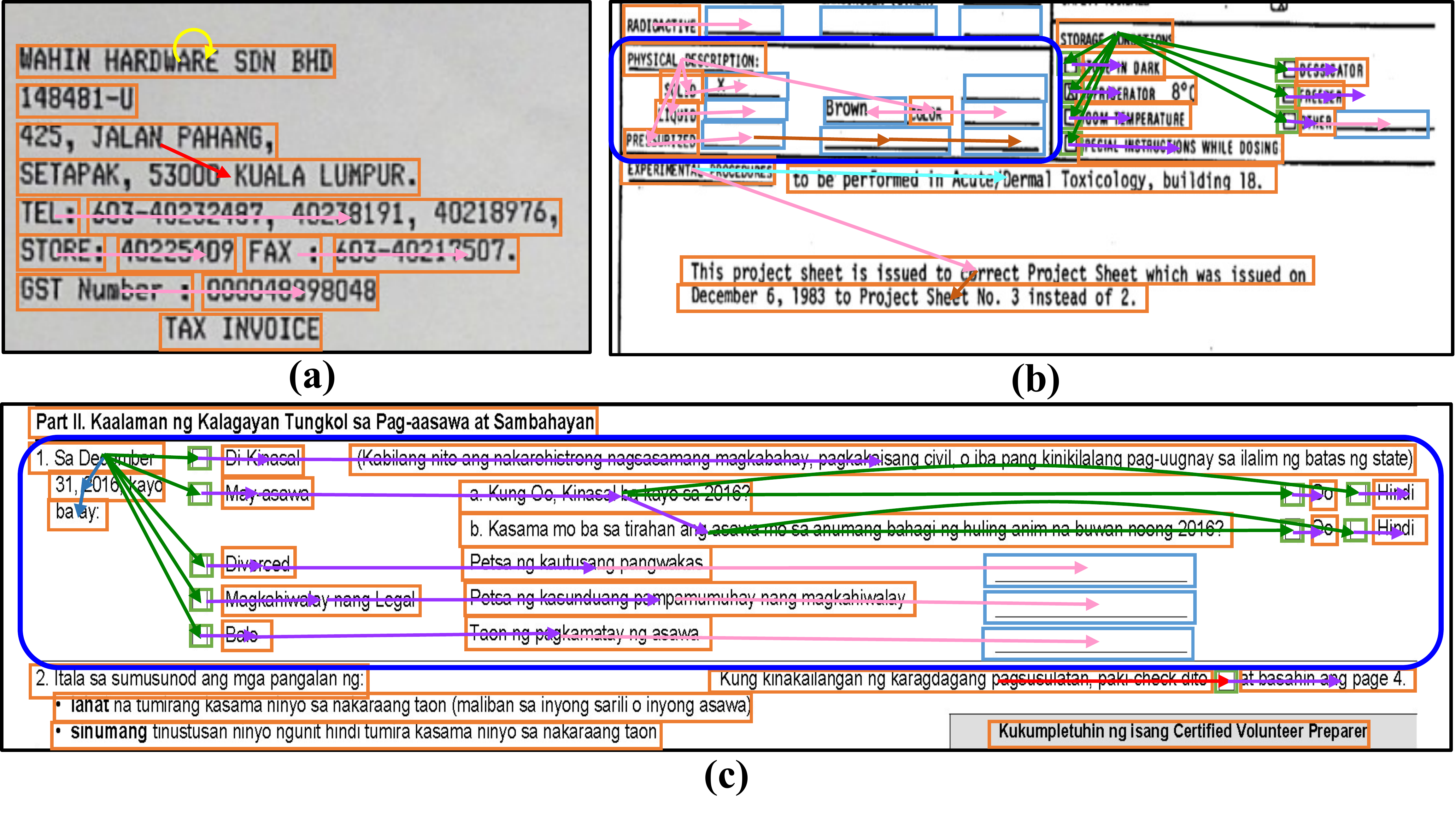}
    \vspace{-3.0mm}
    \caption{An example of our unified label space for Visual Information Extraction: (a) named entities; (b) key-value pairs; (c) choice groups. (Yellow arrow: \textit{intra-company-name}; red arrow: \textit{intra-address}; shy blue arrow: \textit{intra-key}; orange arrow: \textit{intra-value}; pink arrow: \textit{inter-kvp}; blue arrow: \textit{intra-cgt}; purple arrow: \textit{intra-cf}; green arrow: \textit{inter-cg}; blue rectangle: hierarchical key-value pair and choice group. Best viewed in color.)}
    \label{fig:rel-example}
\end{figure}

Given a form-like document image $D$, we identify three fundamental form elements: text-line, text widget, and choice widget. We further delineate two kinds of relationships: \textit{intra-field} and \textit{inter-field} relationships. The objective of these relationships is to group basic form elements into higher-order, structured information units such as named entities, key-value pairs, and choice groups, while preserving their inherent hierarchical structure. Specifically, we define the relationships as follows:
\begin{itemize}
\item As depicted in Fig. \ref{fig:rel-example}(a), for each named entity type denoted as A, we establish an \textit{intra-A} relationship to group text-lines into fields. For fields that comprise a single text-line, we designate the relationship of this text-line as self-referential.
\item As illustrated in Fig. \ref{fig:rel-example}(b), for key-value pairs, we construct \textit{intra-key} and \textit{intra-value} relationships to group basic units into key fields and value fields, respectively. Additionally, we introduce an \textit{inter-kvp} relationship, representing the first basic unit of the key field pointing to the first basic unit of the value field, to reconstruct key-value pairs.
\item As shown in Fig. \ref{fig:rel-example}(c), for choice groups, we define \textit{intra-cgt} and \textit{intra-cf} relationships to group basic units into choice group titles and choice fields, respectively. We also define an \textit{inter-cg} relationship to reconstruct choice groups.
\end{itemize}

These relationships are characterized between basic units, sharing the same relational label space. Importantly, both key-value pairs and choice groups may exhibit a hierarchical structure, potentially encompassing nested relationships as illustrated by the blue rectangle in Fig. \ref{fig:rel-example}(b) and (c). The complexity inherent in such hierarchical structures presents considerable challenges for visual information extraction, necessitating more sophisticated decoding algorithms and deeper semantic analysis to efficiently mine information from these hierarchical structures. Notably, these relationships adhere to a tree-structured pattern, mirroring the hierarchical nature of the information contained within form-like documents.

\section{Methodology}

\begin{figure}[t]
    \centering
    \includegraphics[width=1.0\textwidth]{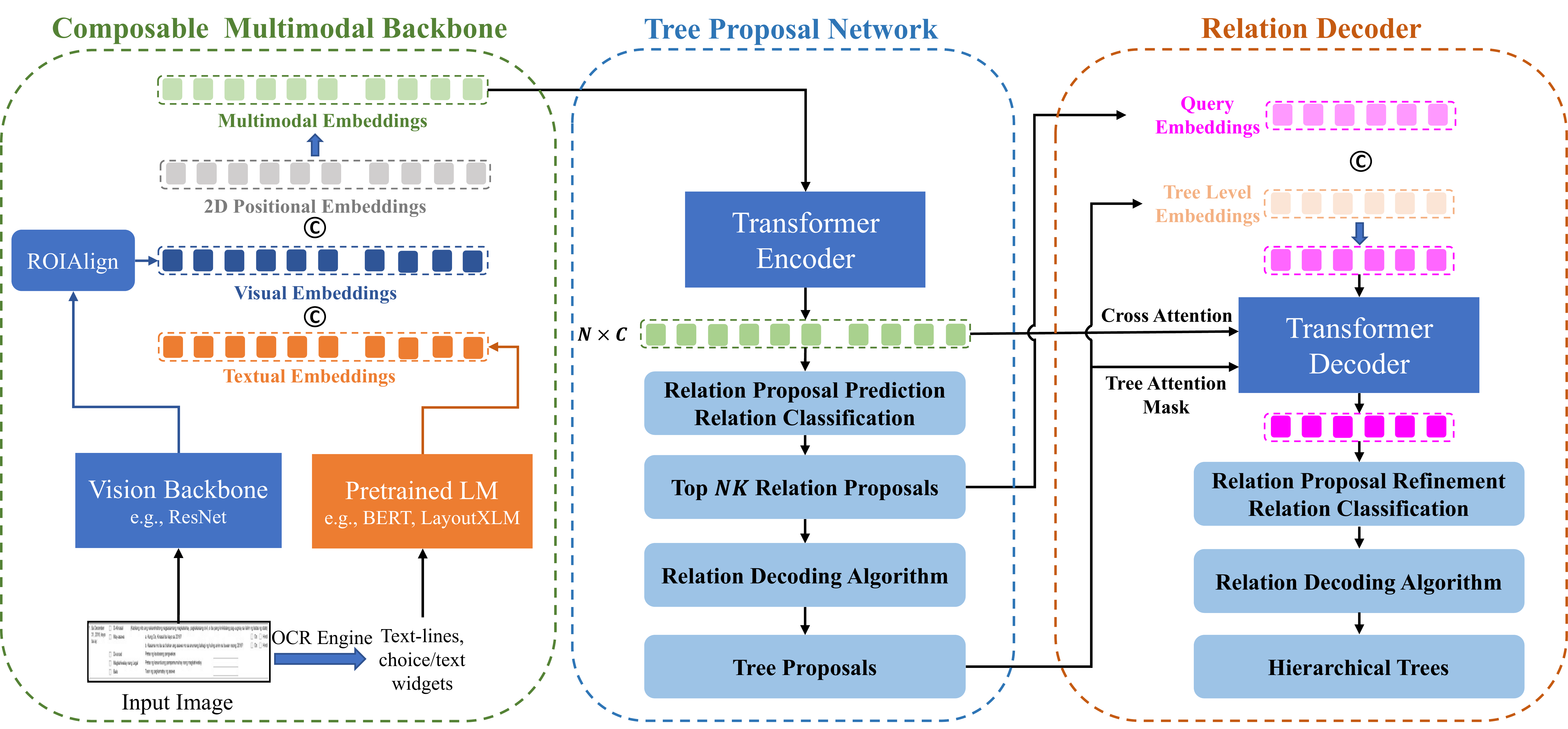}
    \caption{Overview of \Ours{} for Visual Information Extraction.}
    \label{fig:overall}
\end{figure}

In this paper, we cast visual information extraction as a relation prediction problem by introducing a unified label space to unify all pre-defined relationships in a single framework. Building on this formulation, we propose \Ours{} as an example as illustrated in Fig.~\ref{fig:overall}. The architecture of \Ours{} incorporates three core components: (1) A composable multimodal backbone specifically designed for the extraction of multimodal features from form images, (2) A tree proposal network based on a transformer encoder to generate tree proposals, and (3) A relation decoder, which employs the relation proposals as queries to refine these proposals, resulting in the extraction of precise relations. Using these predicted relations, we present a novel relation decoding algorithm, which translates the scored relationships into a set of distinct, non-overlapping hierarchical trees, which represent the structured outcome of our method, capturing hierarchical key-value pairs or choice groups within the form. In following sections, we delve into the details of our \Ours{}.


\subsection{Composable Multimodal Backbone}
Initially, for each input form image, we utilize an OCR engine to extract the bounding box and textual content of all basic units. \Ours{} framework integrates both visual and textual modalities for feature extraction. The visual backbone employs ResNet \cite{he2016deep} to extract visual features from the form image. For each basic unit $t_i$, we apply the RoIAlign algorithm \cite{he2017mask} to extract $7\times7$ features from $C_{3}$, based on its bounding box $b_i = (x_{i1},y_{i1},x_{i2},y_{i2})$, where $(x_{i1},y_{i1})$ and $(x_{i2},y_{i2})$ represent the coordinates of its top-left and bottom-right corners, respectively. Textual features are extracted using pre-trained language models such as Bert \cite{devlin2018bert} or LayoutXLM \cite{xu2021layoutxlm} to derive the text embedding of each basic unit. Specifically, all the basic units in a form image are serialized into a 1D sequence through a top-left to bottom-right reading order, and the basic unit sequence is tokenized into a sub-word token sequence. This token sequence is then fed into the pre-trained language model to yield the embedding of each token. Subsequently, the embeddings of all tokens within each basic unit are averaged to derive its text embedding. Furthermore, we use the bounding box of each basic unit to generate its 2D positional embedding, facilitated by a Multilayer Perceptron layer. The concatenation of these features results in a multimodal embedding for each basic unit.

\subsection{Tree Proposal Network}
\label{sec:tree_proposal}
After obtaining the multimodal embeddings of all basic units, \Ours{} initially employs a Transformer encoder to enhance these embeddings by capturing their interactions via a self-attention mechanism. Each basic unit is treated as an individual token within the Transformer encoder, with its multimodal representation serving as the input embedding. Following this, \Ours{} utilizes these enhanced embeddings to generate tree proposals.

\subsubsection{Relation Proposal Prediction Head.} 
We propose to reframe Visual Information Extraction as a relation prediction task. As depicted in Fig.~\ref{fig:rel-example}, all relationships within our definition adhere to a tree-structured format. 
Inspired by the dependency parsing task \cite{dozat2016deep,zhang2021entity}, which reconstructs the entire dependency parsing tree by predicting the parent word of each word in a sequence, we similarly predict all relationships by identifying the parent basic unit for each basic unit. If a basic unit lacks a parent, it is assigned as its own parent. Assuming a form image consists of $N$ basic units, and two basic units $t_i$ and $t_j$ exist such that $t_i$ points toward $t_j$, our relation prediction head would predict $t_i$ as $t_j$'s parent. Specifically, we employ a multi-class (i.e., $N$-class) classifier to calculate a score $s_{ij}$, estimating the probability of $t_i$ being the parent basic unit of $t_j$ as follows:
\begin{gather}
\label{relation_score}
f_{ij} = MLP(FC_q(F_i) \oplus FC_k(F_j)), \\
s_{ij} = \frac{\exp(f_{ij})}{\sum_{i=1}^{N} \exp(f_{ij})},
\end{gather}
where each of $FC_q$ and $FC_k$ represents a single fully-connected layer with 1,024 nodes, serving to map $F_i$ and $F_j$ into distinct feature spaces; $\oplus$ denotes concatenation operation; MLP consists of 2 fully-connected layers with 1,024 nodes and 1 node respectively. We select the top-$K$ scores from scores $\{s_{kj}, k=1,2,...,n \}$ and output the corresponding basic units as the proposal parents of $t_j$. Consequently, we can obtain a total of $NK$ relation proposals which will be utilized in the subsequent relation decoder module as relation queries to refine more accurate relationships.


\subsubsection{Relation Classification Head.} Once the $NK$ relation proposals are obtained, the ensuing step involves classifying these proposals based on their relation type. To achieve this, we employ a multi-class (i.e., $C$-class) classifier to calculate the relation classification score, where $C$ represents the total count of relation types as per our definition.

\subsubsection{Relation Decoding Algorithm.} 
Following the relation prediction and classification heads, we construct an $N \times N$ relation score matrix, which encapsulates potential connections among the elements of the form. 
Using the relation score matrix, we can construct a weighted fully connected directed graph with each node having a self-loop that points back to itself. To overcome the constraints imposed by self-loops, we construct a virtual node and reassign the scores of all self-loops to point towards this virtual node, thus yielding a modified graph. Our objective is to generate an arborescence (i.e., directed rooted tree) from this modified directed graph, taking the virtual node as the root, while maximizing the total score of the arborescence. This task is the directed analog of the maximum spanning tree problem, for which a classical solution is Chu–Liu/Edmonds' algorithm \cite{chu1965shortest,edmonds1967optimum}. We employ this algorithm to design a relation decoding algorithm, as detailed in Algorithm~\ref{alg:decoding}. This algorithm constructs a maximum spanning tree and subsequently parses out multiple subtrees rooted at the offspring of the virtual node. These subtrees constitute the structured outcome of our algorithm, capturing hierarchical key-value pairs or choice groups within the form. These trees will subsequently be utilized within the relation decoder module as tree constraints in the attention mechanism, thereby facilitating a more effective decoding of refined relationships.

\begin{algorithm}[t]
\caption{Relation Decoding Algorithm}
\label{alg:decoding}
\begin{algorithmic}[1]
\REQUIRE Relation Score Matrix $R \in \mathbb{R}^{N \times N}$, Relation Type Matrix $C \in \mathbb{Z}^{N \times N}$
\STATE Construct a directed complete graph $G$ based on matrices $R$ and $C$.
\STATE Introduce a virtual node, remove all self-loops from $G$, and redirect the self-loops together with their scores to target this virtual node.
\STATE Apply Chu–Liu/Edmonds' algorithm to produce a directed rooted tree $T$ from directed graph $G$, using score matrix $R$ and the virtual node to serve as the root.
\STATE Traverse tree $T$ to identify $K$ subtrees $\{T_i | i = 1, 2, 3, \ldots, K\}$, each rooted at the offspring of the virtual node.
\FOR{$i=1$ \TO $K$}
\STATE Determine the logical role of each node in the subtree $T_i$ based on the types of connecting edges.
\STATE Construct semantically coherent page objects as intermediate nodes within the subtree, using \textit{intra-field} relationships.
\STATE Establish edges between semantically coherent page objects based on \textit{inter-field} relationships.
\STATE Update the hierarchical structure of subtree $T_i$.
\ENDFOR
\RETURN Hierarchical trees $T = \{T_i | i = 1, 2, 3, \ldots, K\}$
\end{algorithmic}
\end{algorithm}


\subsection{Relation Decoder}

\begin{figure}[t]
    \centering
    \includegraphics[width=0.9\textwidth]{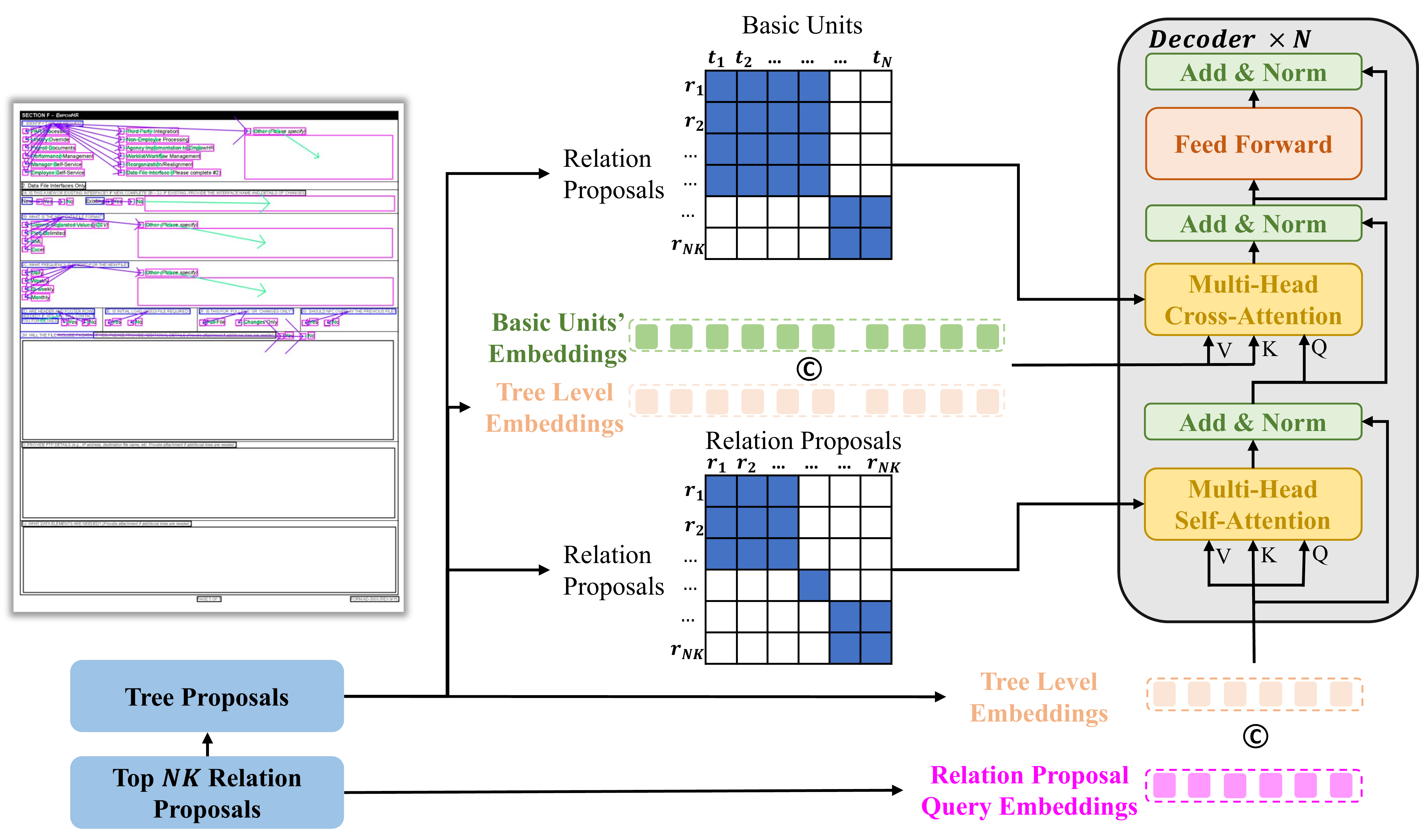}
    \caption{A schematic view of the proposed Relation Decoder module.}
    \label{fig:relation_decoder}
\end{figure}

Departing from previous question-answering based methodologies such as KVPFormer \cite{hu2023question}, which employs a key field as the query and utilizes a transformer decoder to decode the corresponding value field, these techniques are found lacking in their capacity to effectively model the interaction among relations nestled within a hierarchical key-value pair. 
To better capture the interactions among relation proposals and between relation proposals and basic units, as illustrated in Fig.~\ref{fig:relation_decoder}, \Ours{} utilizes relation proposals as queries, employing self-attention to model the interaction among these relation proposals, and cross-attention to model the interaction between these relation proposals and basic units. Furthermore, we introduce two tree constraints within the relation decoder to enhance the effectiveness of our model. 

\subsubsection{Tree Level Embeddings.}
Capitalizing on the previously generated tree proposals, we establish the levels of each basic unit and relation proposal within their respective trees. This level information is then encoded using an embedding layer. The resultant level embeddings are subsequently concatenated to both the query and context embeddings, effectively enriching the feature representation of the basic units and relation proposals.

\subsubsection{Tree Attention Mask.}
To more effectively direct the decoder's attention towards the interactions within each hierarchical tree, we introduce a tree attention mask. This mask operates under a fundamental principle allowing attention flow between relation proposals and basic units within the same tree, while inhibiting attention between entities belonging to different tree proposals. However, it is important to note that the tree proposals are constructed using the top-$1$ parent, while our relation proposals contain top-$K$ potential parents. For those relation proposals that correspond to non-top scoring parents and cross two trees, the potential edges are permitted to attend to the relation proposals and basic units of both trees to provide a broader receptive field for these potential edges.

Upon completion of the relation decoder, we obtain updated relation proposal embeddings. Utilizing these embeddings, we first pass through a relation proposal refinement head. The structure of this head is consistent with the relation proposal prediction head described in Section \ref{sec:tree_proposal}, with the only distinction lying in their operational mechanisms. While the relation proposal prediction head extracts the top-$K$ potential parents from all basic units, the relation proposal refinement head determines the final parent from these top-$K$ potential parents. Subsequent to this, a relation classification head is used to determine the type of these finalized relationships. By applying Algorithm \ref{alg:decoding}, we derive the final hierarchical trees, which constitute the ultimate output of our model.

\subsection{Loss Function}

Within the framework of our \Ours{}, we incorporate two prediction heads into both the \emph{Tree Proposal Network} and the \emph{Relation Decoder} modules. These are the relation prediction head and the relation classification head, each strategically situated within their respective modules. All of these heads consistently employ a softmax cross-entropy as their loss function for effective optimization. The overall loss of our model is determined by aggregating the individual losses from each prediction head.

%% file: chapters/experiments.tex
\section{Experiments}

\subsection{Datasets and Evaluation Protocols}
We conduct experiments on a publicly available dataset (SIBR \cite{yang2023modeling}) and our in-house dataset HierForms to validate the effectiveness of \Ours{}.

\textbf{SIBR}. The SIBR dataset is a public dataset designed for visual information extraction, encompassing 1,000 form images, inclusive of 600 Chinese invoices, 300 English bills of entry, and 100 bilingual receipts. The dataset is divided into a training set of 600 images and a test set of 400 images. There are two predefined tasks in SIBR, entity extraction (EE, \textit{i.e.}, KIE) and entity linking (EL, \textit{i.e.}, KVP). F1-score is employed as an evaluation metric for both EE and EL tasks.

\textbf{HierForms}. For the task of choice group extraction, Adobe researchers introduced a dataset, Forms \cite{aggarwal2021form2seq}, which solely defined single-level choice groups. Unfortunately, due to legal restrictions, only 300 forms were released, which is insufficient for the training of deep learning models. To address this shortcoming, we collect an in-house form dataset, HierForms, which includes human-annotated hierarchical key-value pairs and hierarchical choice groups, significantly more complex than their single-level counterparts. HierForms encompasses 4,488 real-world English form images for training and 500 English form images for testing. We define four types of semantically coherent page objects: choice group title, choice field, key field, and value field, as well as two types of semantically structured page objects: choice groups and key-value pairs. The statistics of this dataset are summarized in Table~\ref{tab:statistics}. We employ the field-level F1-score as the evaluation metric for semantically coherent page objects and tree-level F1-score for semantically structured page objects. The F1-score, being the most stringent evaluation metric, necessitates the accurate prediction of all relationships within a field or tree. Given that semantically structured page objects are tree-structured, we also propose an evaluation metric based on Tree Edit Distance Similarity (TEDS) to assess the tree similarity between the predicted tree and the ground truth tree.

\begin{table}[t]  
\caption{Statistics of HierForms dataset. (CGT: choice group title; CF: choice field; Key: key field; Value: value field; CG: choice group; KVP: key-value pair.)}
\begin{tabularx}{\textwidth}{YYcYYYYcYYcYY}
\toprule
\multirow{3}{*}{Dataset} & \multirow{3}{*}{Images} && \multicolumn{4}{c}{Semantically Coherent} && \multicolumn{5}{c}{Semantically Structured}  \\   \cline{4-7} \cline{9-13}
 & && \multirow{2}{*}{CGT} & \multirow{2}{*}{CF} & \multirow{2}{*}{Key} & \multirow{2}{*}{Value} && \multicolumn{2}{c}{Single Level} && \multicolumn{2}{c}{Hierarchical} \\ \cline{9-10} \cline{12-13}
 & &&  &  &  &  && CG & KVP && CG & KVP \\
 \midrule
 train & 4,488 && 7,598 & 26,702 & 43,956 & 43,980 && 9,922 & 41,032 && 876 & 2,957 \\
 test & 500 && 919 & 3,252 & 5,605 & 5,607 && 1,266 & 5,257 && 185 & 655 \\
 \bottomrule
\end{tabularx}    
\label{tab:statistics}  
\end{table}  

\subsection{Implementation Details}
Our methodology is implemented using PyTorch, and the experiments are conducted on a workstation equipped with 8 NVIDIA Tesla V100 GPUs (32 GB memory). During training, the parameters of the pre-trained language model and visual backbone are initialized with LayoutXLM and ResNet-18, respectively. The transformer encoder of the tree proposal network and the transformer decoder in the relation decoder module are both configured with 3 layers. Both are designed with the number of heads, the dimension of the hidden state, and the dimension of the feedforward network set as 12, 768, and 2048, respectively. The number of relation proposals,  denoted as $K$, is set to 5. The models are optimized using the Adam algorithm \cite{kingma2014adam}, with the learning rate, betas, epsilon, and weight decay configured as 2e-5, (0.9, 0.999), 1e-8 and 1e-2, respectively. All models are trained for 50 epochs with a warmup strategy applied during the first epoch. In each training iteration, a mini-batch of 32 hard positive and 32 hard negative relationships is sampled using the Online Hard Example Mining (OHEM) algorithm \cite{shrivastava2016training}. We employ a multi-scale training strategy, randomly rescaling the shorter side of each image to a value from \{320, 416, 512, 608, 704\}, while ensuring that the longer side does not exceed 800. During inference, the shorter side of each testing image is set to 512.

\subsection{Comparisons with Prior Arts}
In this section, we compare \Ours{} with several state-of-the-art methods on SIBR and HierForms.

\begin{table}[t]  
\centering  
\footnotesize
\caption{Performance comparison on SIBR. (EE: Exitiy Extraction. EL: Entity Linking. The $\dagger$ denotes a co-training process with text spotting branch. The recognition of text is considered correct when the normalized edit distance is less than 0.3.)}  

\begin{tabularx}{0.8\textwidth}{YYYY}  
\toprule
Methods & OCR & EE &  EL  \\
\midrule
TRIE \cite{zhang2020trie} & GT & 85.62 & - \\  
LayoutXLM \cite{xu2021layoutxlm} & GT & 94.72 & 83.99 \\
ESP \cite{yang2023modeling} & GT & 95.27 & 85.96 \\
\Ours{} & GT & \textbf{96.68} & \textbf{87.72} \\
\midrule
TRIE \cite{zhang2020trie} & Duguang & - & - \\
LayoutXLM \cite{xu2021layoutxlm} & Duguang & 68.55 & 46.71 \\
ESP$^\dagger$ \cite{yang2023modeling} & None & 70.45 & 51.47 \\
\Ours{} & Azure Layout & \textbf{83.93} & \textbf{62.83} \\
 \bottomrule
\end{tabularx}  
\label{tab:SIBR}  
\end{table}  

\subsubsection{SIBR.} The SIBR dataset provides the ground truth of text boxes and contents, thereby facilitating the evaluation of \Ours{} under two distinct conditions: (a) with the ground truth of text boxes and contents, and (b) with OCR-processed text boxes and contents. Table~\ref{tab:SIBR} illustrates the performance comparison of \Ours{} with several state-of-the-art methods on the SIBR dataset. Under the first condition (a), where the ground truth of text boxes and contents is provided, \Ours{} surpasses other methods with an Entity Extraction F1-score of 96.68\% and an Entity Linking F1-score of 87.72\%, highlighting its superior ability in accurately extracting and linking entities. For experiments under the second condition (b), we employ Microsoft Azure Layout API\footnote{https://learn.microsoft.com/en-us/azure/ai-services/document-intelligence/concept-layout?view=doc-intel-3.0.0} to obtain word-level locations and contents and use \Ours{} to predict \textit{intra-link} and \textit{inter-link} relationships between OCR-processed words to generate entities and entity linkings. As depicted in Table~\ref{tab:SIBR}, ESP \cite{yang2023modeling} has re-implemented both LayoutXLM \cite{xu2021layoutxlm} and TRIE \cite{zhang2020trie} as baselines, utilizing the Duguang OCR engine\footnote{https://duguang.aliyun.com/} to retrieve character-level locations and content inputs. By jointly training a text spotting branch, ESP has significantly outperformed these baselines. Nonetheless, our proposed model, \Ours{}, still exhibits superior performance, achieving an Entity Extraction F1-score of 83.93\% and an Entity Linking F1-score of 62.83\%, both of which significantly surpass those of the other methods.


\begin{table}[t]  
\caption{Performance comparison on HierForms. (CGT: choice group title, CF: choice field, Key: key field, Value: value field, CG: choice group, KVP: key-value pair.)}  
\begin{tabularx}{\textwidth}{cc|ccYYYYcYYcYY}
\toprule
\multirow{3}{*}{Input} & \multirow{3}{*}{Level} & \multirow{3}{*}{Methods} && \multicolumn{4}{c}{Semantically Coherent} && \multicolumn{5}{c}{Semantically Structured}  \\   \cline{5-8} \cline{10-14}
 && && \multicolumn{4}{c}{F1-score} && \multicolumn{2}{c}{F1-score} && \multicolumn{2}{c}{TEDS} \\ \cline{5-8} \cline{10-11} \cline{13-14}
 && && CGT & CF & Key & Value && CG & KVP && CG & KVP \\
 \midrule
\multirow{6}{*}{GT} & \multirow{3}{*}{Single} & LayoutLMv2 \cite{xu2020layoutlmv2} && 78.7 & 70.0 & 88.1 & 82.9 && 55.4 & 72.9 && 64.2 & 75.8 \\
&& KVPFormer \cite{hu2023question} && 82.1 & 82.7 & 92.0 & 90.6 && 63.2 & 86.7 && 74.0 & 88.2 \\
&& \Ours{} && \textbf{83.4} & \textbf{84.8} & \textbf{92.4} & \textbf{91.3} && \textbf{65.2} & \textbf{88.3} && \textbf{78.4} & \textbf{90.5} \\ \cline{2-14}
& \multirow{3}{*}{\shortstack{Single \& \\ Hierarchical}} & LayoutLMv2 \cite{xu2020layoutlmv2} && 75.8 & 75.2 & 82.1 & 82.2 && 47.6 & 63.2 && 51.5 & 65.5 \\
&& KVPFormer \cite{hu2023question} && 77.8 & 80.6 & 84.2 & 83.2 && 58.6 & 72.5 && 63.2 & 75.2 \\
&& \Ours{} && \textbf{84.3} & \textbf{88.1} & \textbf{91.5} & \textbf{91.1} && \textbf{64.8} & \textbf{76.9} && \textbf{70.4} & \textbf{80.5} \\
 \midrule
 \midrule
\multirow{6}{*}{OCR} & \multirow{3}{*}{Single} & LayoutLMv2 \cite{xu2020layoutlmv2} && 76.4 & 69.8 & 83.8 & 80.9 && 52.1 & 70.8 && 57.8 & 68.6 \\
&& KVPFormer \cite{hu2023question} && 79.2 & 81.3 & 85.4 & 88.5 && 56.3 & 79.3 && 62.3 & 81.5 \\
&& \Ours{} && \textbf{80.0} & \textbf{84.1} & \textbf{86.6} & \textbf{88.8} && \textbf{62.8} & \textbf{81.3} && \textbf{72.9} & \textbf{84.2} \\ \cline{2-14}
& \multirow{3}{*}{\shortstack{Single \& \\ Hierarchical}} & LayoutLMv2 \cite{xu2020layoutlmv2} && 74.9 & 68.3 & 81.9 & 78.8 && 43.6 & 58.2 && 48.4 & 61.9 \\
&& KVPFormer \cite{hu2023question} && 75.4 & 78.0 & 83.5 & 82.0 && 55.1 & 67.0 && 61.2 & 71.5 \\
&& \Ours{} && \textbf{80.6} & \textbf{85.6} & \textbf{85.7} & \textbf{88.3} && \textbf{61.7} & \textbf{69.8} && \textbf{68.9} & \textbf{73.4} \\
 \bottomrule
\end{tabularx}  
\label{tab:HierForms}  
\end{table}  

\subsubsection{HierForms.} 
The HierForms dataset comes with ground truth annotations limited to text bounding boxes at the line level. To capture the useful textual information, we employ Microsoft Azure Layout API to extract transcripts along with their corresponding word-level bounding boxes from the form images. These enriched annotations enable us to conduct a comprehensive evaluation of our proposed \Ours{}, under two distinct experimental conditions: (a) using the ground truth text-line bounding boxes as input, and (b) using the OCR-processed word bounding boxes as input.

We implemented two baseline methods to address both the single level and hierarchical extraction of key-value pairs and choice groups. The first method is an extension of LayoutLMv2 \cite{xu2020layoutlmv2}, wherein an additional Multi-Layer Perceptron (MLP) relation classification layer was incorporated to predict pre-defined relationships. The second approach utilized KVPFormer \cite{hu2023question}, a method based on a question-answering paradigm. By formulating each key field or choice group title as a question, KVPFormer leverages a QA mechanism to infer the corresponding value fields and choice fields.

As summarized in Table~\ref{tab:HierForms}, under all evaluation conditions, our proposed \Ours{} significantly outperforms the two baseline approaches for single level extraction. It achieves a Tree Edit Distance-based Similarity (TEDS) score of 78.4\% for choice groups and 90.5\% for key-value pairs when utilizing ground truth text-line bounding boxes as input. When employing OCR-processed word bounding boxes as input, \Ours{} remains the best with a TEDS score of 72.9\% for choice groups and 84.2\% for key-value pairs, respectively. When extended to tackle the more challenging hierarchical extraction scenarios, \Ours{} consistently outperforms both LayoutLMv2 and KVPFormer. Specifically, when employing ground truth text-line bounding boxes as input, \Ours{} achieves a TEDS score that outperforms KVPFormer by a substantial 7.2\% for choice groups and by an impressive 5.3\% for key-value pairs. Moreover, when utilizing OCR-processed word bounding boxes as input, \Ours{} extends its lead, outperforming KVPFormer by 7.7\% for choice groups and by 1.9\% for key-value pairs. These results highlight \Ours{}'s exceptional ability to interpret and extract information from forms, even in the presence of inherent OCR errors.


\subsection{Ablation Studies}

In this section, we conduct a series of ablation studies to assess the contributions of each component within \Ours{}. All experiments are performed on the HierForms dataset due to its larger size.

\begin{table}[t]  
\caption{Ablation studies of different component in \Ours{} on HierForms dataset, where ``TLE" means Tree Level Embeddings and ``TAM" means Tree Attention Mask. (CG: choice group, KVP: key-value pair.)}  
\begin{tabularx}{\textwidth}{YcYYcYYcYYc|YY}
\toprule
 \multirow{2}{*}{\#} && \multicolumn{2}{c}{Multimodal} && \multicolumn{2}{c}{Architecture} && \multicolumn{2}{c}{Tree Constraints} && \multicolumn{2}{c}{TEDS} \\ \cline{3-4} \cline{6-7} \cline{9-10} \cline{12-13}
 && Image & Text && Encoder & Decoder && TLE &  TAM && CG & KVP \\ 
 \midrule
 \midrule
  \Ours{} &&  &  &&  &  &&  &  && \textbf{70.4} & \textbf{80.5} \\
  \midrule
 1a && \xmark & &&  &  &&  &  && 68.9 & 79.3 \\
 1b &&  & \xmark  &&  &  &&  &  && 64.0 & 74.3 \\
 \midrule
 2a &&  &  && \xmark  &  &&  &  && 66.4 & 78.8 \\
 2b &&  &  &&  & \xmark &&  &  && 65.2 & 76.9 \\
 2c &&  &  && \xmark & \xmark &&  &  && 55.1 & 74.3 \\
 \midrule
 3a &&  &  &&  &  && \xmark &  && 69.5 & 79.3 \\
 3b &&  &  &&  &  && \xmark & \xmark && 67.3 & 78.5 \\
 \bottomrule
\end{tabularx}  
\label{tab:ablation}  
\end{table}  

\subsubsection{Effectiveness of Multimodal Features.} 
The contribution of each modality to \Ours{}'s performance is evaluated by individually excluding image and text modalities. As illustrated in the rows 1a and 1b of Table~\ref{tab:ablation}, the exclusion of either image or text modalities leads to a decline in performance. A more notable degradation in performance is seen when the text modality is removed, suggesting that textual information is more crucial than visual information for the VIE task. The late fusion of both visual and textual information outperforms the use of either visual or textual information in isolation.

\subsubsection{Effect of Transformer Encoder-Decoder Architecture.}
The results presented in the rows 2a, 2b, and 2c of Table~\ref{tab:ablation} demonstrate that 1) Both the encoder and decoder play pivotal roles in improving performance; 2) Decoder-only models slightly surpass encoder-only models; 3) The combination of both encoder and decoder results in optimal performance.

\subsubsection{Impact of Tree Constraints.}
The rows 3a and 3b of Table~\ref{tab:ablation} assess the significance of our proposed tree constraints, namely Tree Level Embeddings (TLE) and Tree Attention Mask (TAM). The degradation in performance observed on the removal of TLE (row 3a), and a further decline when both TLE and TAM are excluded (row 3b), underscores the substantial influence of these tree constraints on the model's performance.

%% file: chapters/conclusion.tex
\section{Conclusion and Future Work}

In this paper, we propose \Ours{}, a unified model designed for Visual Information Extraction (VIE) from form-like documents. This model provides a new perspective, reframing VIE as a relation prediction task and consequently unifying labels across diverse tasks into a single label space. \Ours{} employs a coarse-to-fine strategy, initially generating tree proposals through a tree proposal network, and subsequently refining them into hierarchical trees via a relation decoder. Moreover, we introduce two novel tree constraints, a tree attention mask and a tree level embedding, to bolster the relation prediction capabilities of the relation decoder. These novel additions significantly enhance \Ours{}'s overall effectiveness. Comprehensive experimental evaluations underscore \Ours{}'s superiority over existing methodologies. The promising results from this study indicate that label unification and relation prediction are promising avenues for future research in VIE field. 

In the future, we will continue to investigate the capabilities of our method utilizing a unified label space in the context of zero-shot and few-shot visual information extraction scenarios. The exploration is expected to illuminate the adaptability and robustness of our approach, particularly in scenarios where training data is scarce or entirely absent.

%% file: main.bbl
\begin{thebibliography}{10}
\providecommand{\url}[1]{\texttt{#1}}
\providecommand{\urlprefix}{URL }
\providecommand{\doi}[1]{https://doi.org/#1}

\bibitem{aggarwal2021form2seq}
Aggarwal, M., Gupta, H., Sarkar, M., Krishnamurthy, B.: Form2seq: A framework for higher-order form structure extraction. In: Proceedings of the Conference on Empirical Methods in Natural Language Processing (EMNLP). pp. 3830--3840 (2020)

\bibitem{aggarwal2020multi}
Aggarwal, M., Sarkar, M., Gupta, H., Krishnamurthy, B.: Multi-modal association based grouping for form structure extraction. In: Proceedings of the IEEE/CVF Winter Conference on Applications of Computer Vision (WACV). pp. 2075--2084 (2020)

\bibitem{carbonell2021named}
Carbonell, M., Riba, P., Villegas, M., Forn{\'e}s, A., Llad{\'o}s, J.: Named entity recognition and relation extraction with graph neural networks in semi structured documents. In: Proceedings of the International Conference on Pattern Recognition (ICPR). pp. 9622--9627 (2021)

\bibitem{cesarini2003analysis}
Cesarini, F., Francesconi, E., Gori, M., Soda, G.: Analysis and understanding of multi-class invoices. Document Analysis and Recognition  \textbf{6},  102--114 (2003)

\bibitem{chu1965shortest}
Chu, Y.J.: On the shortest arborescence of a directed graph. Scientia Sinica  \textbf{14},  1396--1400 (1965)

\bibitem{dengel2002smartfix}
Dengel, A.R., Klein, B.: smartfix: A requirements-driven system for document analysis and understanding. In: Proceedings of Document Analysis Systems (DAS). pp. 433--444 (2002)

\bibitem{devlin2018bert}
Devlin, J., Chang, M.W., Lee, K., Toutanova, K.: Bert: Pre-training of deep bidirectional transformers for language understanding. In: Proceedings of the Annual Conference of the North American Chapter of the Association for Computational Linguistics: Human Language Technologies (NAACL-HLT). pp. 4171--4186 (2019)

\bibitem{dozat2016deep}
Dozat, T., Manning, C.D.: Deep biaffine attention for neural dependency parsing. In: Proceedings of the International Conference on Learning Representations (ICLR) (2017)

\bibitem{edmonds1967optimum}
Edmonds, J., et~al.: Optimum branchings. Journal of Research of the national Bureau of Standards B  \textbf{71}(4),  233--240 (1967)

\bibitem{esser2012automatic}
Esser, D., Schuster, D., Muthmann, K., Berger, M., Schill, A.: Automatic indexing of scanned documents: a layout-based approach. In: Proceedings of Document Recognition and Retrieval (DRR). pp. 118--125 (2012)

\bibitem{gao2021value}
Gao, M., Xue, L., Ramaiah, C., Xing, C., Xu, R., Xiong, C.: Docquerynet: Value retrieval with arbitrary queries for form-like documents. In: Proceedings of the International Conference on Computational Linguistics (COLING). pp. 2141--–2146 (2022)

\bibitem{gupta2016table}
Gupta, P., Sch{\"u}tze, H., Andrassy, B.: Table filling multi-task recurrent neural network for joint entity and relation extraction. In: Proceedings of the International Conference on Computational Linguistics (COLING). pp. 2537--2547 (2016)

\bibitem{he2017mask}
He, K., Gkioxari, G., Doll{\'a}r, P., Girshick, R.: Mask r-cnn. In: Proceedings of the IEEE International Conference on Computer Vision (WACV). pp. 2961--2969 (2017)

\bibitem{he2016deep}
He, K., Zhang, X., Ren, S., Sun, J.: Deep residual learning for image recognition. In: Proceedings of the IEEE Conference on Computer Vision and Pattern Recognition (CVPR). pp. 770--778 (2016)

\bibitem{hu2023question}
Hu, K., Wu, Z., Zhong, Z., Lin, W., Sun, L., Huo, Q.: A question-answering approach to key value pair extraction from form-like document images. Proceedings of the AAAI Conference on Artificial Intelligence (AAAI) pp. 12899--12906 (2023)

\bibitem{huang2019icdar2019}
Huang, Z., Chen, K., He, J., Bai, X., Karatzas, D., Lu, S., Jawahar, C.: Icdar2019 competition on scanned receipt ocr and information extraction. In: Proceedings of International Conference on Document Analysis and Recognition (ICDAR). pp. 1516--1520 (2019)

\bibitem{hwang2020spatial}
Hwang, W., Yim, J., Park, S., Yang, S., Seo, M.: Spatial dependency parsing for semi-structured document information extraction. In: Findings of the Association for Computational Linguistics (ACL). pp. 330--–343 (2021)

\bibitem{jaume2019funsd}
Jaume, G., Ekenel, H.K., Thiran, J.P.: Funsd: A dataset for form understanding in noisy scanned documents. In: Proceedings of International Conference on Document Analysis and Recognition Workshops (ICDARW). pp.~1--6 (2019)

\bibitem{kingma2014adam}
Kingma, D.P., Ba, J.: Adam: A method for stochastic optimization. In: Proceedings of the International Conference on Learning Representations (ICLR) (2015)

\bibitem{lin2021vibertgrid}
Lin, W., Gao, Q., Sun, L., Zhong, Z., Hu, K., Ren, Q., Huo, Q.: Vibertgrid: A jointly trained multi-modal 2d document representation for key information extraction from documents. In: Proceedings of the International Conference on Document Analysis and Recognition (ICDAR). pp. 548--563 (2021)

\bibitem{mathur2023layerdoc}
Mathur, P., Jain, R., Mehra, A., Gu, J., Dernoncourt, F., Tran, Q., Kaynig-Fittkau, V., Nenkova, A., Manocha, D., Morariu, V.I., et~al.: Layerdoc: Layer-wise extraction of spatial hierarchical structure in visually-rich documents. In: Proceedings of the IEEE/CVF Winter Conference on Applications of Computer Vision (WACV). pp. 3610--3620 (2023)

\bibitem{medvet2011probabilistic}
Medvet, E., Bartoli, A., Davanzo, G.: A probabilistic approach to printed document understanding. International Journal on Document Analysis and Recognition (IJDAR)  \textbf{14},  335--347 (2011)

\bibitem{park2019cord}
Park, S., Shin, S., Lee, B., Lee, J., Surh, J., Seo, M., Lee, H.: Cord: A consolidated receipt dataset for post-ocr parsing. In: Document Intelligence Workshop at Neural Information Processing Systems (2019)

\bibitem{qiao2022joint}
Qiao, B., Zou, Z., Huang, Y., Fang, K., Zhu, X., Chen, Y.: A joint model for entity and relation extraction based on bert. Neural Computing and Applications pp. 1--11 (2022)

\bibitem{rastogi2020information}
Rastogi, M., Ali, S.A., Rawat, M., Vig, L., Agarwal, P., Shroff, G., Srinivasan, A.: Information extraction from document images via fca based template detection and knowledge graph rule induction. In: Proceedings of the IEEE/CVF Conference on Computer Vision and Pattern Recognition Workshops (CVPRW). pp. 558--559 (2020)

\bibitem{rusinol2013field}
Rusinol, M., Benkhelfallah, T., Poulain~dAndecy, V.: Field extraction from administrative documents by incremental structural templates. In: Proceedings of the International Conference on Document Analysis and Recognition (ICDAR). pp. 1100--1104 (2013)

\bibitem{schuster2013intellix}
Schuster, D., Muthmann, K., Esser, D., Schill, A., Berger, M., Weidling, C., Aliyev, K., Hofmeier, A.: Intellix -- end-user trained information extraction for document archiving. In: Proceedings of the International Conference on Document Analysis and Recognition (ICDAR). pp. 101--105 (2013)

\bibitem{shrivastava2016training}
Shrivastava, A., Gupta, A., Girshick, R.: Training region-based object detectors with online hard example mining. In: Proceedings of the IEEE Conference on Computer Vision and Pattern Recognition (CVPR). pp. 761--769 (2016)

\bibitem{vsimsa2023docile}
{\v{S}}imsa, {\v{S}}., {\v{S}}ulc, M., U{\v{r}}i{\v{c}}{\'a}{\v{r}}, M., Patel, Y., Hamdi, A., Koci{\'a}n, M., Skalick{\'y}, M., Matas, J., Doucet, A., Coustaty, M., Karatzas, D.: Docile benchmark for document information localization and extraction. In: Proceedings of the International Conference on Document Analysis and Recognition (ICDAR). pp. 147--166 (2023)

\bibitem{wang-etal-2022-lilt}
Wang, J., Jin, L., Ding, K.: Lilt: A simple yet effective language-independent layout transformer for structured document understanding. In: Proceedings of the Annual Meeting of the Association for Computational Linguistics (ACL). pp. 7747--7757 (2022)

\bibitem{wang2020two}
Wang, J., Lu, W.: Two are better than one: Joint entity and relation extraction with table-sequence encoders. In: Proceedings of the Conference on Empirical Methods in Natural Language Processing (EMNLP). pp. 1706--1721 (2020)

\bibitem{wang2021unire}
Wang, Y., Sun, C., Wu, Y., Zhou, H., Li, L., Yan, J.: Unire: A unified label space for entity relation extraction. In: Proceedings of the Annual Meeting of the Association for Computational Linguistics and the International Joint Conference on Natural Language Processing (ACL-IJCNLP). pp. 220--–231 (2021)

\bibitem{wang2020tplinker}
Wang, Y., Yu, B., Zhang, Y., Liu, T., Zhu, H., Sun, L.: Tplinker: Single-stage joint extraction of entities and relations through token pair linking. In: Proceedings of the International Conference on Computational Linguistics (COLING). pp. 1572–--1582 (2020)

\bibitem{watanabe1995layout}
Watanabe, T., Luo, Q., Sugie, N.: Layout recognition of multi-kinds of table-form documents. IEEE Transactions on Pattern Analysis and Machine Intelligence  \textbf{17}(4),  432--445 (1995)

\bibitem{xu2020layoutlmv2}
Xu, Y., Xu, Y., Lv, T., Cui, L., Wei, F., Wang, G., Lu, Y., Florencio, D., Zhang, C., Che, W., et~al.: Layoutlmv2: Multi-modal pre-training for visually-rich document understanding. In: Proceedings of the Annual Meeting of the Association for Computational Linguistics and the International Joint Conference on Natural Language Processing (ACL-IJCNLP). pp. 2579--–2591 (2021)

\bibitem{xu2020layoutlm}
Xu, Y., Li, M., Cui, L., Huang, S., Wei, F., Zhou, M.: Layoutlm: Pre-training of text and layout for document image understanding. In: Proceedings of the ACM SIGKDD International Conference on Knowledge Discovery \& Data Mining. pp. 1192--1200 (2020)

\bibitem{xu2021layoutxlm}
Xu, Y., Lv, T., Cui, L., Wang, G., Lu, Y., Florencio, D., Zhang, C., Wei, F.: Xfund: A benchmark dataset for multilingual visually rich form understanding. In: Findings of the Association for Computational Linguistics (ACL). pp. 3214--3224 (2022)

\bibitem{yan2022utc}
Yan, H., Sun, Y., Li, X., Zhou, Y., Huang, X., Qiu, X.: Utc-ie: A unified token-pair classification architecture for information extraction. In: Proceedings of the Annual Meeting of the Association for Computational Linguistics (ACL). pp. 4096--4122 (2023)

\bibitem{yang2023modeling}
Yang, Z., Long, R., Wang, P., Song, S., Zhong, H., Cheng, W., Bai, X., Yao, C.: Modeling entities as semantic points for visual information extraction in the wild. In: Proceedings of the IEEE/CVF Conference on Computer Vision and Pattern Recognition (CVPR). pp. 15358--15367 (2023)

\bibitem{zhang2020trie}
Zhang, P., Xu, Y., Cheng, Z., Pu, S., Lu, J., Qiao, L., Niu, Y., Wu, F.: Trie: End-to-end text reading and information extraction for document understanding. In: Proceedings of the ACM International Conference on Multimedia. pp. 1413--1422 (2020)

\bibitem{zhang2021entity}
Zhang, Y., Bo, Z., Wang, R., Cao, J., Li, C., Bao, Z.: Entity relation extraction as dependency parsing in visually rich documents. In: Proceedings of the Conference on Empirical Methods in Natural Language Processing (EMNLP). pp. 2759--2768 (2021)

\bibitem{zheng2017joint}
Zheng, S., Hao, Y., Lu, D., Bao, H., Xu, J., Hao, H., Xu, B.: Joint entity and relation extraction based on a hybrid neural network. Neurocomputing  \textbf{257},  59--66 (2017)

\end{thebibliography}
